\documentclass[utf8]{FrontiersinHarvard} % for articles in journals using the Harvard Referencing Style (Author-Date), for Frontiers Reference Styles by Journal: https://zendesk.frontiersin.org/hc/en-us/articles/360017860337-Frontiers-Reference-Styles-by-Journal
\usepackage{url,hyperref,microtype,subcaption}
\usepackage[onehalfspacing]{setspace}
\usepackage{mathptmx}
\usepackage{amsfonts}
\usepackage{amsmath}
\usepackage{booktabs}
\newcommand{\beginsupplement}{
  \setcounter{table}{0}  
  \renewcommand{\thetable}{S\arabic{table}} 
  \setcounter{figure}{0} 
  \renewcommand{\thefigure}{S\arabic{figure}}
}
\def\keyFont{\fontsize{8}{11}\helveticabold }
\def\firstAuthorLast{Powadi {et~al.}} %use et al only if is more than 1 author
\def\Authors{Anirudha Powadi\,$^{1}$, Talukder Zaki Jubery,\,$^{2}$, Michael C. Tross, \,$^{3,5}$, James C. Schnable, \,$^{3,5,*}$ and Baskar Ganapathysubramanian, \,$^{1,2,4}$}
\def\Address{$^{1}$Department of Electrical and Computer Engineering, Iowa State University, Ames, 5011, IA, USA \\
$^{2}$ Translational AI Research and Education Center, Iowa State University, Ames, 50011, Ames, USA \\
$^{3}$ Department of Agronomy and Horticulture, University of Nebraska-Lincoln, Lincoln, 68588, Nebraska, USA \\ 
$^{4}$ Department of Mechanical Engineering, Iowa State University, Ames, 50011, Iowa, USA \\
$^{5}$ Center for Plant Science Innovation, University of Nebraska-Lincoln, Lincoln, 68588, Nebraska, USA}
\def\corrAuthor{2529 Union Dr, Ames, IA 50011}

\def\corrEmail{baskarg@iastate.edu}

\begin{document}
\onecolumn
\firstpage{1}

\title[CAE]{Disentangling Genotype and Environment Specific Latent Features for Improved Trait Prediction using a Compositional Autoencoder} 

\author[\firstAuthorLast ]{\Authors} %This field will be automatically populated
\address{} %This field will be automatically populated
\correspondance{} %This field will be automatically populated

\extraAuth{}% If there are more than 1 corresponding author, comment this line and uncomment the next one.
%\extraAuth{corresponding Author2 \\ Laboratory X2, Institute X2, Department X2, Organization X2, Street X2, City X2 , State XX2 (only USA, Canada and Australia), Zip Code2, X2 Country X2, email2@uni2.edu}

\maketitle
\begin{abstract}

This study introduces a compositional autoencoder (CAE) framework designed to disentangle the complex interplay between genotypic and environmental factors in high-dimensional phenotype data to improve trait prediction in plant breeding and genetics programs.  
Traditional predictive methods, which use compact representations of high-dimensional data through handcrafted features or latent features like PCA or more recently autoencoders, do not separate genotype-specific and environment-specific factors. We hypothesize that disentangling these features into genotype-specific and environment-specific components can enhance predictive models. To test this, we developed a compositional autoencoder (CAE) that decomposes high-dimensional data into distinct genotype-specific and environment-specific latent features.

Our CAE framework employs a hierarchical architecture within an autoencoder to effectively separate these entangled latent features. Applied to a maize diversity panel dataset, the CAE demonstrates superior modeling of environmental influences and 5-10 times improved predictive performance for key traits like Days to Pollen and Yield, compared to the traditional methods, including standard autoencoders, PCA with regression, and Partial Least Squares Regression (PLSR). By disentangling latent features, the CAE provides powerful tool for precision breeding and genetic research. This work significantly enhances trait prediction models, advancing agricultural and biological sciences. 

\tiny
 \keyFont{ \section{Keywords:} Hierarchical Disentanglement, Latent Disentanglement, Plant Phenotyping} %All article types: you may provide up to 8 keywords; at least 5 are mandatory.
\end{abstract}

\section{Introduction}
\subsection{Background and Overview}
Advances in imaging and robotic technologies are making both high-resolution images and sensor data increasingly accessible to plant biologists and breeders as tools to capture measurements of plant traits. These data types can be used to measure or predict traits which are labor intensive or costly to measure directly including variation in plant architectural and biochemical traits as well as resistance or susceptibility to specific biotic stresses.  A growing body of evidence suggests high dimensional trait datasets can also be useful to predict crop productivity (e.g. grain yield)~\citep{adak2023temporal,yieldpreds_measurements}.  However, like the plant traits plant biologists and breeders seek to predict, sensor data and the high dimensional traits extracted from that data reflect the impact of both genetic and environmental factors. Traditionally, the data are used in raw form or through handcrafted and engineered features without explicitly decomposing the genotype (G) and environment (E) factors. The very high dimensionality of sensor data often makes hand-crafting features a very non-trivial and challenging task. Under this 'curse of dimensionality', hand-crafted and engineered features can fail to capture the full variability within the dataset, potentially limiting accuracy and/or impairing the ability of trained models to translate to new environments or new sets of plant genotypes. Latent features derived from these data are better than hand-crafted or engineered features because they can capture the underlying patterns and variability in the dataset without being biased by human assumptions or limited by predefined feature sets ~\citep{LatentSpacePhenoReview,aguate2017use}. This can lead to more accurate and generalizable models that can better predict complex traits and their interactions.

Although sensors and images capture a great deal of information about a target plant or target plants, not all of that information is likely to be useful. Latent phenotyping is an emerging approach to plant phenotyping that seeks to minimize human bias in defining differences between plants by reducing the dimensionality of the data via unsupervised or self-supervised approaches ~\citep{LatentSpacePheno1,LatentSpacePheno2,LatentSpacePhenoReview,MichaelHyperspectral}. Traditionally, machine learning methods like PCA (Principal component analysis), Linear Discriminant Analysis (LDA), T-distributed Stochastic Neighbor Embedding (t-SNE), and autoencoders have been used to extract the `latent representation' from high-dimensional data ~\citep{PCA_LDA,DataRepresentationLearning,LatentReprBio,ImageReprPCA,VarAEMetabolomics,LatentGalaxySpectra}. A specific advantage of autoencoders, relative to many of the other methods employed to extract latent representations from sensor or image data is their ability to capture non-linear relationships within data as well as being extremely configurable. An autoencoder has two parts. The first part works to produce a compressed but highly informative latent space, while the second part works to rebuild the original input from this latent space. This process enables the autoencoder to effectively learn a compact, yet robust, representation of the input data, capturing essential features and patterns in a lower-dimensional latent space, which is crucial for accurate data reconstruction.

Dimensionality reduction techniques endeavor to produce a compressed but highly informative `latent representation' from the input data. This latent representation is often used to predict or classify some quantity of interest related to the input data. This representation although very informative, is not categorized. This can be explained by a simple example of generating a latent space of an image of a corn plant. Let us say that our goal is for the latent space to effectively capture and quantify various external attributes such as `number of leaves', `plant height', `chlorophyll concentration (greenness)', and `position of the corn cob'. However, a typical challenge arises in this process: the elements within the latent space tend to be intermixed or `entangled.' This means that instead of having distinct areas in the latent space dedicated to each specific attribute (like one area for ear position, another for leaf number, etc.), these characteristics are all woven together. As a result, the latent space does not neatly separate one feature from another but rather blends them in a complex and interconnected way. Disentangling this latent space can often result in significant improvements in downstream prediction accuracy and has been a topic of significant research effort in recent years. 

\subsection{Related Work and Contributions}
There are several ways to achieve disentanglement, but disentanglement often comes at the cost of reconstruction accuracy. One approach is via regularization techniques, where additional terms are added to the loss function (of, say, vanilla and variational autoencoders, VAE) \citep{Kingma_2019} to encourage independence among latent variables. Examples include $\beta$-VAE \citep{higgins2017betavae}, which balances reconstruction fidelity and disentanglement. FactorVAEs \citep{Dis-Factor} impose a total correlation penalty on the latent variables to promote their independence and disentanglement. Mutual information-based approaches, like InfoGAN and StyleGAN, maximize the mutual information between latent variables and generated outputs to ensure distinct factors of variation are captured. Additionally, supervised or semi-supervised approaches can be employed, leveraging labeled data to guide the learning of disentangled representations ~\citep{supervisedDisentanglement,semiSupervisedGenerativeModel,kingma2022autoencodingvariationalbayes}. Latent feature disentangling has been successfully applied to a diverse array of applications including music ~\citep{music}, text understanding ~\citep{text}, facial image generation ~\citep{StyleGAN}, protein structure variations~\citep{proteinAE}. 

Disentanglement can be broadly classified into two types: hierarchical disentanglement and latent space disentanglement. Hierarchical disentanglement organizes the latent space into multiple levels, where higher levels capture abstract and global features, and lower levels capture specific and local details. On the other hand, latent space disentanglement ensures that each dimension in the latent space corresponds to a distinct and independent factor of variation, promoting independence among latent variables. Both approaches aim to create more interpretable and modular representations of data, each with a different focus on the structure and independence of the latent features. Approaches like $\beta$-Variational autoencoder ~\citep{burgess2018understanding,higgins2017betavae} and its variants~\citep{guerrerolópez2022multimodal,8953731,watters2019spatial,orthogonality} perform latent space disentanglement -- disentangling features that control size, position and orientation of an input image. Similarly, StyleGAN and its variations ~\citep{Liu2022-ur,Niu2023-em,10223056} start directly with a latent vector picked from a Gaussian distribution, and associate specific features of the image with individual components of the latent vector.  Hierarchical disentanglement has been richly explored on applications involving speech ~\citep{speech}, video sequences ~\citep{videoSequences}, multi-modal temporal data ~\citep{multi-temporal}. These approaches use concepts of attention ~\citep{attention}, context addition ~\citep{context}, graph convolution ~\citep{graphC}, or contrastive learning ~\citep{contrastive}. Although these papers have shown exciting results, a limitation is the inability to simultaneously utilize multiple sources of data to perform hierarchical disentanglement. This issue is circumvented in approaches like Orthogonal denoising autoencoder ~\citep{OrthoAE} and Factorized latent space paper ~\citep{FactorLatentSpace} that disentangle while learning the features presented by side views of an object. Another approach ~\citep{OrthoFacialRep} works on disentanglement via enforcing a correlation loss that penalizes correlation between identity and facial expression representation. These two concepts were leveraged in our work along with the concept of latent disentanglement. 

%\subsection{Problem Definition and Contribution}
High-dimensional sensor data have been recently used for plant phenotyping due to their capability to quantify plant features in an unbiased fashion while scaling to larger experiments. Autoencoders have been used to extract a concise representation from this high-dimensional data while also filtering out noise \citep{LatentSpacePheno1,LatentSpacePheno2,MichaelHyperspectral}. This extracted representation comes with certain drawbacks. First, it is not interpretable, i.e, we do not know what information is exactly embedded into it. Second, it does not have any structure or hierarchy to it. Finally, it does not encode different factors of variation in the data separately. To tackle these problems, we employ a combination of hierarchical and latent disentanglement strategies.

\begin{figure}[ht!]
    \centering
    \includegraphics[width=1\textwidth]{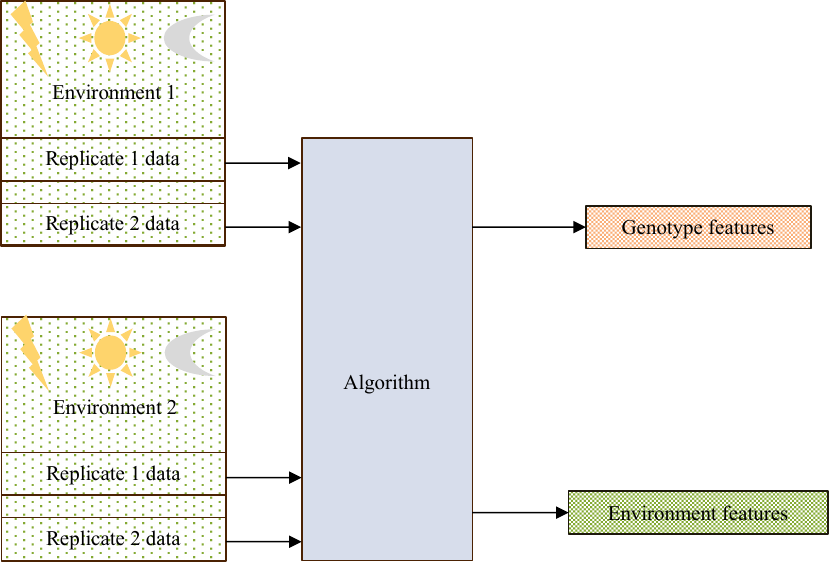}
 \caption{The problem definition: Extract and disentangle the effects of genotype and environment for a given type of sensor data, assuming multiple observations of each genotype in each environment. Our specific test dataset was a set of hyperspectral reflectance data collected from 578 distinct genotypes of maize in two distinct environments, with two replicates of each genotype in each environment (four total replicates per genotype and 1,156 total replicates per environment).}
 \label{fig:probfig}
\end{figure}

Our goal in this paper is to disentangle abstract and hierarchical effects which are not apparent from the data. In particular, we seek strategies towards disentangling the effects of the environment (latent variables that represent the effect of soil conditions, weather, water, and treatment) and genotype (latent variables that represent the effect of genes or genetic makeup) when compressing and reconstructing high dimensional data (specifically, hyperspectral data) that represent phenotypic measurements of maize plants. We believe that disentangling the latent space into environment and gene effects should help improve the predictive performance of the learnt representation on many downstream tasks. Figure.~\ref{fig:probfig} schematically illustrates the problem definition.

Our contributions in this paper are as follows:
    a) we report a generalized architecture -- compositional autoencoder (CAE) -- that can produce a disentangled, low-dimensional, latent representation (that respects hierarchical relationships), given high-dimensional data across a diverse set of plant genotypes. In this case, the effects of genotype and environment on hyperspectral data collected from plants. 
    b) This architecture (CAE) shows an improvement in predicting `Days to Pollen', a measure of flowering time which plays a key role in determining crop variety suitability to different environments, when compared to standard vanilla autoencoder or PCA. 
    c) The CAE latent representation produces models with improved accuracy in predicting the trait `Yield' (i.e. the amount of grain produced by a given crop variety grown on a fixed amount of land), which is both critically important to farmers and considered quite difficult to predict from mid-season sensor measurements, when compared to the current state-of-art methods like classical autoencoders. 
    
\begin{figure}[ht]
    \centering
    \includegraphics[width=1\textwidth]{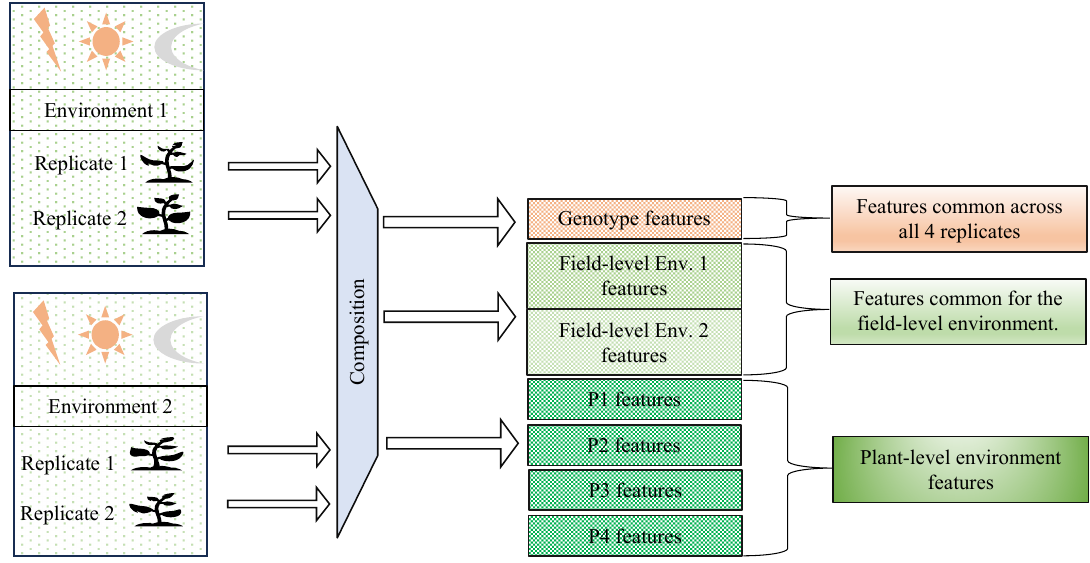}
 \caption{The goal of this research is to disentangle the input hyperspectral data into genotype-specific information, environment specific information and plant-specific information. This achieved by the method of composition.}
 \label{fig:Workflow}
\end{figure}

Figure \ref{fig:Workflow} shows the visual representation of the disentangled latent space.
The effects of the environment could be anything from the weather, soil conditions, to the treatments administered to plants in a field. 

\begin{figure}[ht]
    \centering
    \includegraphics[width=1\textwidth]{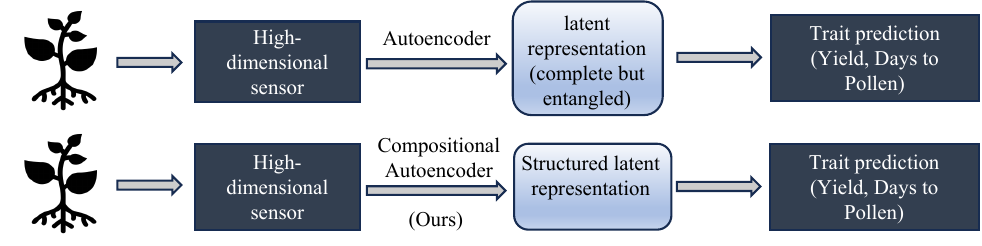}
 \caption{Trait prediction workflow of a Vanilla Autoencoder vs Compositional Autonencoder.}
 \label{fig:Overview}
\end{figure}
The figure \ref{fig:Overview} shows an overview of the current approaches versus our (CAE) approach.

\section{Materials and Methods}

\subsection{Equipment and Dataset}

\begin{figure}[h]
    \centering
    \includegraphics[width=1\textwidth]{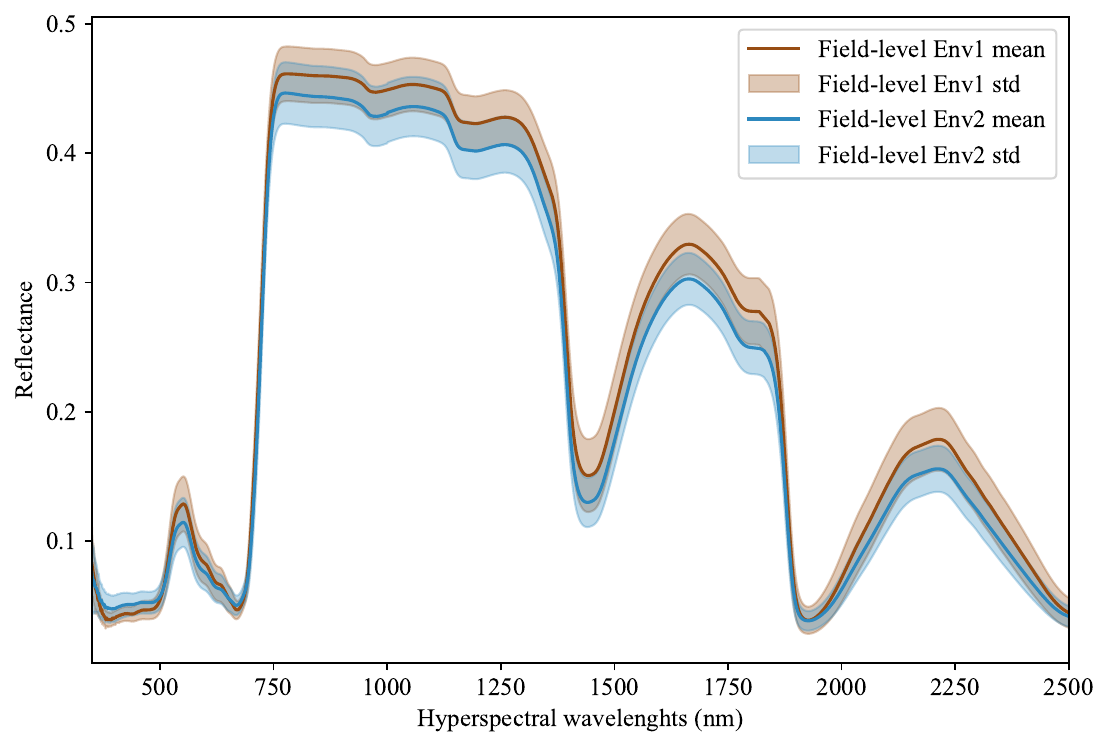}
    \caption{Hyperspectral leaf reflectance data was collected using a FieldSpec4 (Malvern Panalytical Ltd., Formerly Analytical Spectral Devices) with a contact probe. A total of 2151 wavelengths were collected, ranging from 350 nm to 2500 nm. The dataset consists of measurements for a set of 578 different maize inbred genotypes that were grown and phenotyped in two different environments with 2 replicates per environment.}
    \label{fig:Dataset}
\end{figure}

Hyperspectral data is being increasingly adopted by plant scientists as a method to measure or predict plant traits in field and greenhouse settings ~\citep{NovelHypPollen,HyperSpecImageVariety,High-ThroughputPhenMaizeRef,MichaelHyperspectral}. For the purposes of this study, we employed data from 578 inbreds, which represent a subset of the Wisconsin Diversity panel \citep{mazaheri2019genome}, grown and phenotyped in 2020 and 2021 at the Havelock Farm research facility at the University of Nebraska-Lincoln. In each year, measurements were collected on two replicated plots of each inbred grown in different parts of the field, for a total $2 \times 2 \times 578 = 2312$ observed plots. Each plot consisted of two rows of genetically identical plants with approximately 20 plants per row, as previously described in \citet{mural2022association}. Hyperspectral data was collected using FieldSpec4 spectroradiometers (Malvern Panalytical Ltd., Formerly Analytical Spectral Devices) with a contact probe. This equipment captures 2151 wavelengths of electromagnetic radiation ranging from 350 nm to 2500 nm. 
Hyperspectral data was collected from a single fully expanded leaf per plot, selected from a representative plant, avoiding edge plants whenever possible. Three spectral measurements were taken at each of the three points located at the tip, middle, and base of the adaxial side of each leaf. Values were averaged across the nine wavelength scans to generate a final composite spectrum for each plot sampled \citep{MichaelHyperspectral}.  Figure \ref{fig:Dataset} illustrates the distribution and variability of mean reflectance among the genotypes across two years, which in this paper are referred to as two different environments. We divide the environment into field-level (or macro-environment) and plot-level (or micro-environment) ~\cite{micro-macro}.
%$value_{norm} = (value - min_{dataset}) / (max_{dataset} - min_{dataset})$
For the latent features extraction, the data was then normalized using min-max normalization. This normalization is given as:
\begin{equation}
x_{\text{normalized}} = \frac{x - min_{\text{dataset}}}{max_{\text{dataset}} - min_{\text{dataset}}}
\label{eq:minmax}
\end{equation}
From the equation \ref{eq:minmax}, '$min_{dataset}$' and '$max_{dataset}$' are the minimum and maximum values in the entire dataset respectively.
\newpage

\subsection{Vanilla Autoencoder}

\begin{figure}[ht!]
    \centering
    \includegraphics[width=0.7\textwidth]{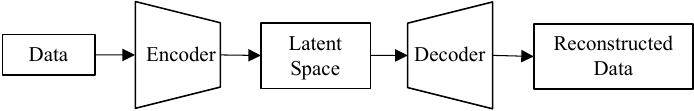}
 \caption{A vanilla autoencoder works to learn a compressed yet highly informative representation of the input data.}
 \label{fig:VanillaAE}
\end{figure}

We implemented a standard autoencoder (see Figure \ref{fig:VanillaAE}) as a baseline for comparison which we refer to below as the `vanilla autoencoder' (AE).
Both the encoder and decoder portions of our vanilla autoencoder implementation are made up of multiple fully connected layers stacked together with the non-linear activation function `SeLu.' The encoder encodes the input data (2151 wavelengths) into smaller dimensions (latent space) and decoder works to reconstruct back the original input from this latent space. The tables \ref{table:encoder} and \ref{table:decoder} show the details of each of the layers that constitute the encoder and decoder.
For training the vanilla autoencoder, data from each plot in each year is considered as one sample, resulting in a total of 2312 input samples.

\begin{table}[ht!]
    \centering
    \caption{Encoder: Configuration Details. `input\_shape' = 1 x 2151, `zg' = dimensions allocated to capture genotype features, `ze' = dimensions allocated to capture macro-environment features, `zp' = dimensions allocated to micro-environment features.}
    \begin{tabular}{ccc}
        \hline
        \textbf{Layer Type} & \textbf{Dimensions} & \textbf{Activation} \\
        \hline
        Linear & \texttt{input\_shape} $\to$ 2150 & SELU \\
        Linear & 2150 $\to$ 1024 & SELU \\
        Linear & 1024 $\to$ 512 & SELU \\
        Linear & 512 $\to$ \texttt{zg + ze + zp} & None \\
        \hline
    \end{tabular}
    \label{table:encoder}
\end{table}

\begin{table}[ht!]
\centering
\caption{Decoder: Configuration Details. `input\_shape' = 1 x 2151, `zg' = dimensions allocated to capture genotype features, `ze' = dimensions allocated to capture macro-environment features, `zp' = dimensions allocated to micro-environment features.}
\begin{tabular}{ccc}
    \hline
    \textbf{Layer Type} & \textbf{Dimensions} & \textbf{Activation} \\
    \hline
    Linear & \texttt{zg + ze + zp} $\to$ 512 & SELU \\
    Linear & 512 $\to$ 1024 & SELU \\
    Linear & 1024 $\to$ 2150 & SELU \\
    Linear & 2150 $\to$ \texttt{input\_shape} & Sigmoid \\
    \hline
\end{tabular}
\label{table:decoder}
\end{table}

\subsection{Compositional Autoencoder}
\subsubsection{Architecture}
The compositional autoencoder extends the vanilla autoencoder architecture in a way that aims to disentangle the latent space, partitioning the impact of different factors that influence the data into different variables. It consists of an encoder, decoder, and a fusion block. The network operates as follows:

\begin{enumerate}
    \item Encode Individual Plant Data: The encoder processes data from four plants of the same genotype, compressing it into latent features.
    \item Fuse Encoded Data: These encoded representations from all the plants are then fused into a single latent feature.
    \item Disentangle Latent Factors: This fused latent feature is then partitioned into three distinct parts: genotype-specific features (common across all plants), macro-environment-specific features (shared by plants from the same environment), and micro-environment-specific features (unique to each plant).
    \item Reconstruct Individual Plants: Finally, for each plant, the genotype, macro-environment, and micro-environment features are assembled. This assembled disentangled representation is then decoded to reconstruct the original plant data.
\end{enumerate}

\begin{figure}[ht]
    \includegraphics[width=1\textwidth]{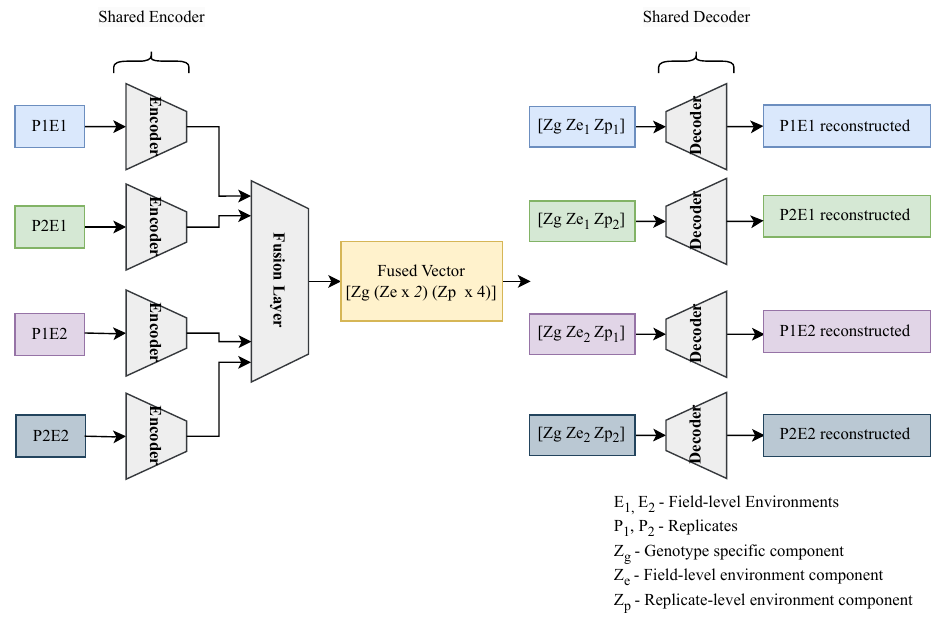}
 \caption{The encoder encodes the hyperspectral data for 4 plants, accounting for a single genotype across two environments (E1, and E2) and 2 replicated per environment (P1E1, P2E1, P1E2, P2E2). The resulting 4 latent vectors are fused using a linear layer. The resulting fused vector contains 3 parts. a) Genotype representation part. b) Macro or field-level environment representation part (2 parts to represent E1 and E2 effects). c) Micro or replicate-specific environment representation component (4 parts to represent each of the plants (P1E1, P2E1, P1E2, P2E2)). To get the composed encoded form, genotype representation is combined with the field-level environment part and plant-level environment part. These composed encoded vectors are then fed into the decoder to regenerate the original hyperspectral reflectance.}
 \label{fig:NN_Arch}
\end{figure}

Here, genotype refers to groups of plants with identical genetic makeups, macro-environment refers to common environmental factors experienced by all plants growing in the same field in the same year (e.g. rainfall, temperature), and micro-environment refers to features of the individual replicate growing in the same field within the same environment/year. The table (refer to Table \ref{tab:plant-rep}) illustrates the disentangled latent representation for each plant. A more detailed network architecture can be found in the figure (refer to Figure \ref{fig:NN_Arch}).
The encoder and decoder used here are the same as vanilla autoencoder with the addition of `Fusion' layer. The layer details are provided in the table \ref{table:encoder_fusion}.
\begin{table}[ht!]
\centering
\caption{Fusion Layer Details. `N' = number of replicates per genotype (2), `E' = number of environments (2), `zg' = dimensions allocated to capture genotype features, `ze' = dimensions allocated to capture macro-environment features, `zp' = dimensions allocated to micro-environment features.}
\begin{tabular}{ccc}
    \hline
    \textbf{Layer Type} & \textbf{Dimensions} & \textbf{Activation} \\
    \hline
    Linear & \texttt{N(zg + ze + zp)} $\to$ \texttt{zg + E(ze) + N(zp)} & None \\
    \hline
\end{tabular}
\label{table:encoder_fusion}
\end{table}

\begin{table}[ht]
    \centering
    \caption{Disentangled latent-space representation of each plant.}
    \begin{tabular}{cc}
        \hline
        \textbf{Plant} & \textbf{Representation} \\
        \hline
        Plant 1 & \{(Zg) genotype, (Ze) macro-environment [1], (Zp) micro-environment [1]\} \\
        Plant 2 & \{(Zg) genotype, (Ze) macro-environment [1], (Zp) micro-environment [2]\} \\
        Plant 3 & \{(Zg) genotype, (Ze) macro-environment [2], (Zp) micro-environment [3]\} \\
        Plant 4 & \{(Zg) genotype, (Ze) macro-environment [2], (Zp) micro-environment [4]\} \\
        \hline
    \end{tabular}
    \label{tab:plant-rep}
\end{table}

The training process involves dividing the hyperspectral data into groups of four plants (sharing the same genotype). There are a total of 578 such groups (corresponding to the number of genotypes). Each group is fed sequentially through the encoder, resulting in four latent representations. These representations are then fused together. The resulting fused latent space captures three factors: genotype, field-level environment (with two sub-parts for the two environments), and plant-level environment (with four sub-parts for the four plants).

\subsubsection{Loss Function}

We trained the CAE network using a two-part loss function consisting of a reconstruction loss and a correlation loss.

\textit{Reconstruction Loss:} The mean squared error (MSE), was used as the reconstruction loss for the compositional autoencoder. This loss function encourages the network to learn a meaningful disentangled latent space that can be accurately decoded back to the original hyperspectral data.

% \begin{equation}
%   \text{MSE} = \frac{1}{n} \sum_{i=1}^{n} (Y_i - \hat{Y}_i)^2
%   \label{eq:mse}
% \end{equation}

% where:

% \begin{itemize}
%     \item $n$ is the number of data points.
%     \item $Y_i$ is the actual (ground truth) value for the $i$-th data point.
%     \item $\hat{Y}_i$ is the predicted value for the $i$-th data point.
% \end{itemize}

\textit{Correlation Loss:} A correlation loss was employed to ensure that all parts in the disentangled latent space remain uncorrelated throughout the training process. This loss is defined in Equation (\ref{eq:CorrLoss}).

\begin{equation}
  \text{Correlation Loss} = \sum_{i=1}^{N} \sum_{j=i}^{N} \left| \text{CorrMat}_{ij} \right| - I_{ij}
  \label{eq:CorrLoss}
\end{equation}

where:

\begin{itemize}
    \item $\text{CorrMat}_{ij}$ represents the correlation coefficient between dimensions $i$ and $j$ in the latent space.
    \item $N$ is the dimension of the square correlation matrix, which corresponds to the number of dimensions in the latent space.
    \item $I_{ij}$ is the identity matrix, ensuring that the diagonal elements (where $i = j$) contribute zero to the loss.
\end{itemize}

The correlation coefficient used here is the Pearson correlation coefficient ($r$), a measure of the linear correlation between two variables. It is calculated using Equation (\ref{eq:pearson}).

\begin{equation}
  r = \frac{\sum_{i=1}^{n} (p_i - \bar{p}) (k_i - \bar{k})}{\sqrt{\sum_{i=1}^{n} (p_i - \bar{p})^2 \sum_{i=1}^{n} (k_i - \bar{k})^2}}
  \label{eq:pearson}
\end{equation}

where:

\begin{itemize}
    \item $n$ is the number of data points.
    \item $p_i$ and $k_i$ are the elements of the latent space.
    \item $\bar{p}$ and $\bar{k}$ are the means of the $p^{th}$ dimension and $k^{th}$ dimension, respectively.
\end{itemize}

In our case, we aim to achieve zero correlation between the latent space features representing genotype, environment, and individual plant variations. This is enforced by the correlation loss function (Equation (\ref{eq:CorrLoss})). This ensures that the disentangled latent space captures these factors independently.

We trained the vanilla autoencoder network using MSE reconstruction loss only.

\subsubsection{Training Parameters}
The data was divided into training and validation with a 85\%-15\% split. Furthermore, we trained these networks with SGD, Adam, and LBFGS optimizers and found that LBFGS gave us faster convergence (10x). Therefore, all the experiments were carried out using the LBFGS optimizer. The training setup included early stopping criteria, which monitored validation loss and stopped training after it observed no improvements in the metric for 15 epochs. 

\subsubsection{Parameter Tuning for Downstream Tasks}
    To improve the performance of latent representations for downstream tasks, we investigated several tuning techniques for both the network and its inputs. 
    \begin{itemize}
        \item a) We explored masking a portion of the input data. This technique encourages the model to focus on reconstructing the missing parts, potentially leading to increased robustness and reduced overfitting \citep{MultiMAE}. We performed a search for the optimal masking percentage.
        \item b) Considering our dataset size, we conducted a basic architecture search to strike a balance between model complexity and data availability. This helps to mitigate overfitting and improve generalization. We evaluated different network architectures with varying numbers of layers and dimensions in the encoder and decoder.
        \item  c)  To ensure the latent representations captured the necessary data complexity, we experimented with different latent space dimensions and their composition of genotype, field-level, and plant-level environmental features.
    \end{itemize}

\subsection{Downstream Tasks Performance Metrics}
 To confirm our hypothesis that the disentangled latent representations enhance the latent feature's ability to predict useful traits, we generated disentangled latent features (disentangled encoded output from the encoder) for all 2312 data points. We then used these features to train models to predict two traits, namely, `Days to Pollen' and `Yield (grams)'. We trained several regression models --- Random Forests, XGBoost, Ridge Regressions, and PLSR (Partial-Least Square Regression) --- to identify a high performing model. We compare the performance of the models trained on the disentangled latent representations from the CAE against the performance of models trained on the latent representations from a vanilla autoencder. The resulting prediction performance was evaluated using an R\textsuperscript{2} metric representing the coefficient of determination.The coefficient of determination, $R^2$, is defined as:

\begin{equation}
    R^2 = 1 - \frac{\sum_{i=1}^{n} (y_i - \hat{y}_i)^2}{\sum_{i=1}^{n} (y_i - \bar{y})^2}
\end{equation}

where:
\begin{itemize}
    \item $y_i$ is the observed value,
    \item $\hat{y}_i$ is the predicted value, and
    \item $\bar{y}$ is the mean of the observed data.
\end{itemize}

\section{Results and Discussion}
\subsection{Disentangled Representation from CAE}
\begin{figure}
    \centering
    \begin{tabular}{cc}
        \includegraphics[width=0.45\textwidth]{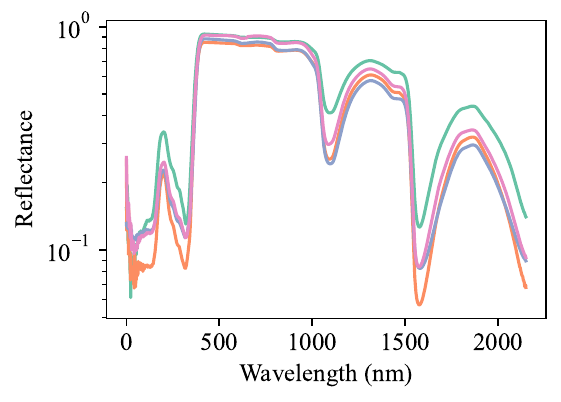} &
        \includegraphics[width=0.45\textwidth]{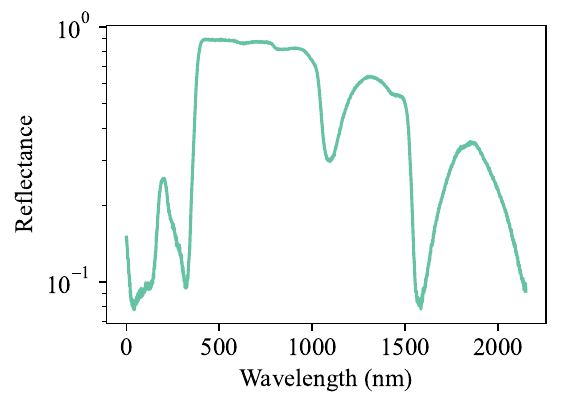} \\
        \parbox[t]{0.45\textwidth}{\centering(a) Measured Reflectance\label{fig:measured_reflectance}} &
        \parbox[t]{0.45\textwidth}{\centering(b) Disentangled Genotype Reflectance\label{fig:disentangled_genotype_reflectance}} \\[1em]
        \includegraphics[width=0.45\textwidth]{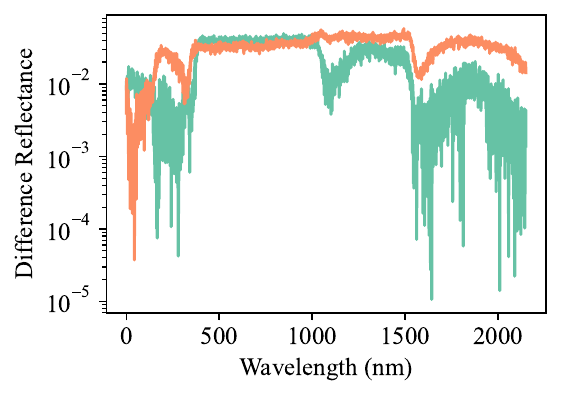} &
        \includegraphics[width=0.45\textwidth]{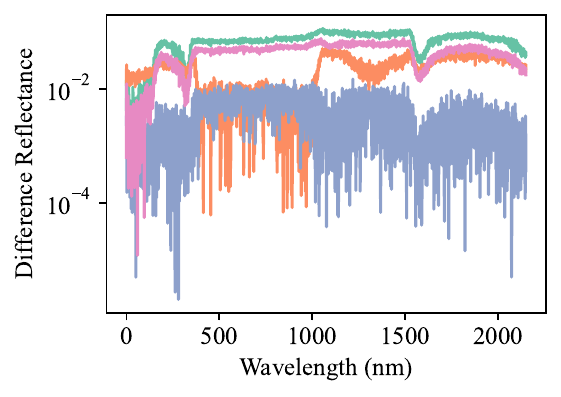} \\
        \parbox[t]{0.45\textwidth}{\centering(c) Disentangled Macro Environmental Influence\label{fig:disentangled_macro_environmental_influence}} &
        \parbox[t]{0.45\textwidth}{\centering(d) Disentangled Micro Environmental Influence\label{fig:disentangled_micro_environmental_influence}}
    \end{tabular}
    \caption{Reflectance Measurements and Disentangled Influences: (a) the original measured reflectance spectra for multiple samples of a particular genotype, and the disentangled reflectance components attributed to (b) genotype, (c) macro-environmental influence, and (d) micro-environmental influence. Note the significant variation in the original reflectance due to the combined effects of genotype and environment. Disentanglement enables the visualization of distinct spectral patterns associated with each factor, highlighting the CAE's ability to separate these influences.}
    \label{fig:Reconstructions}
\end{figure}

The compositional autoencoder (CAE) successfully disentangled the latent space into genotype, macro- and micro- environmental effects. The figure \ref{fig:Reconstructions} shows a comparison of the original reflectance versus factor-specific (genotype and environments) reflectance. Here, factor-specific reflectance is obtained by modifying the latent space to only keep the effects of either the genotype, or the environments; and subsequently reconstructing the reflectance from them. Therefore, genotype-specific is obtained by replacing the environment components in the latent space with an average of all the environments, and similarly, genotype components are replaced by their average to reconstruct the environment-specific reflectance. Figure \ref{fig:Reconstructions}b shows genotype-specific reflectance. As we are focusing on just 1 genotype in this figure, all the replicates will have the same latent space and therefore, the same reflectance. Figure \ref{fig:Reconstructions}c shows macro environment-specific reflectance. The distinction between the two macro-environments is visualized by calculating the difference between macro-environment-specific reflectance and genotype-specific reflectance for the two macro-environments. Similarly, figure \ref{fig:Reconstructions}d shows micro-environment-specific reflectance. The visualization shows the difference between genotype-specific reflectance, macro-environment-specific reflectance, and micro-environment-specific reflectance. 

\begin{figure}[h!]
    \centering
    \begin{tabular}{cc}
        \includegraphics[width=0.49\textwidth]{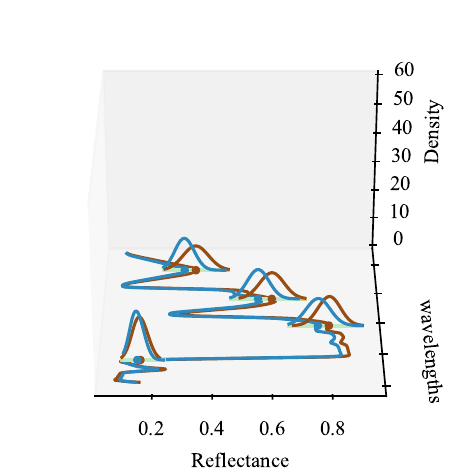} &
        \includegraphics[width=0.49\textwidth]{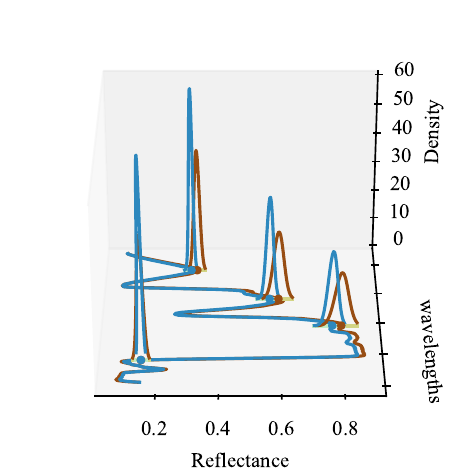} \\
        \parbox[t]{0.49\textwidth}{\centering (a) Original Macro environmental reflectance for the two macro environments.\label{fig:env1_env2}} &
        \parbox[t]{0.49\textwidth}{\centering (b) Disentangled Macro environmental effects\label{fig:env1r_env2r}}
    \end{tabular}
    \caption{This figure shows the disentanglement achieved for the environmental effects using the CAE. Two sets of visualizations are presented: the original environmental effects (Env 1 and Env 2) and the disentangled versions. The average KL-divergence observed for the original input data is \textbf{0.62} while the disentangled KL-divergence is \textbf{2.79}. The density distribution of the reflectance values is shown at a couple of wavelengths instead of all wavelengths for better visualization. This illustrates the separation of environmental factors before and after applying the CAE, indicating the model's effectiveness in disentangling these effects.}
    \label{fig:env_influ}
\end{figure}

To further verify the degree of environment disentanglement, we calculated the distribution of the two macro environments for the original reflectance (Figure \ref{fig:env_influ}a) and disentangled environments' reflectance (Figure \ref{fig:env_influ}b). A successful disentanglement should yield completely separated distributions. We use KL-divergence to measure the difference between the distributions. We can clearly see that KL-divergence of distributions representing two environments generated from the sensor data is quite low (0.62) while the same for the disentangled reflectance is quite large (2.79). This strongly indicates that the latent representation is, in fact, able to represent the two environments distinctly.

\subsection{Performance of Latent Representations on Downstream tasks}

\begin{figure}[h!]
    \centering
    \begin{tabular}{cc}
        \includegraphics[width=0.44\linewidth]{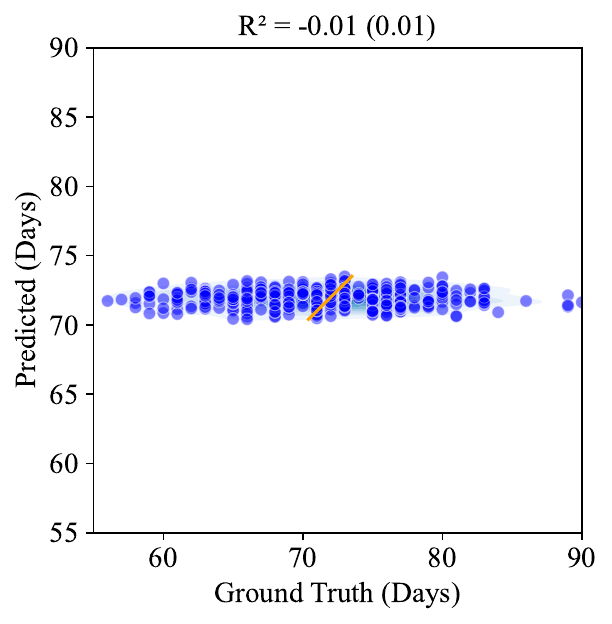} &
        \includegraphics[width=0.47\linewidth]{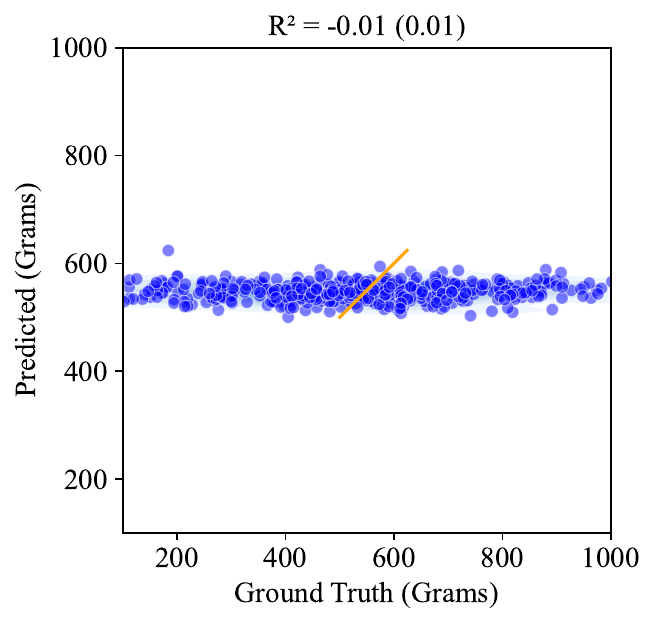} \\
        \parbox[t]{0.44\linewidth}{\centering (a) Performance on `Days to Pollen'\label{fig:VAEDaysToPollen}} &
        \parbox[t]{0.47\linewidth}{\centering (b) Performance on `Yield'\label{fig:VAETotalGrainMassGrams}}
    \end{tabular}
    \caption{Performance of AE on `Days to Pollen' and `Yield' using Ridge regression: (left) Days to Pollen: The x-axis represents ground truth days (60-90 days), and the y-axis represents predicted days (55-90 days). The scatter plot shows points widely scattered, indicating poor prediction accuracy. The model achieves an R² of -0.01 (±0.01), demonstrating negligible correlation between predicted and actual values. (right) Total Grain Mass: The x-axis represents ground truth grain mass (200-1000 grams), and the y-axis represents predicted grain mass (200-1000 grams). The scatter plot shows points widely scattered, indicating poor prediction accuracy. The model achieves an R² of -0.01 (±0.02), demonstrating negligible correlation between predicted and actual values.}
    \label{fig:VAE_Results}
\end{figure}

We first report on the performance of our baseline model -- the vanilla autoencoder. The latent representation from the vanilla AE was used to train a multiple machine learning models to predict the two traits. We present the Ridge regression model performance here as it yielded the best results among all the models (Random Forests, PLSR, and XgBoost). Figure \ref{fig:VAE_Results} shows this performance. We see that the performance for both the traits in question is quite low ($r^2 = ~0.01$).

\begin{figure}[h!]
    \centering
    \begin{tabular}{cc}
        \includegraphics[width=0.45\linewidth]{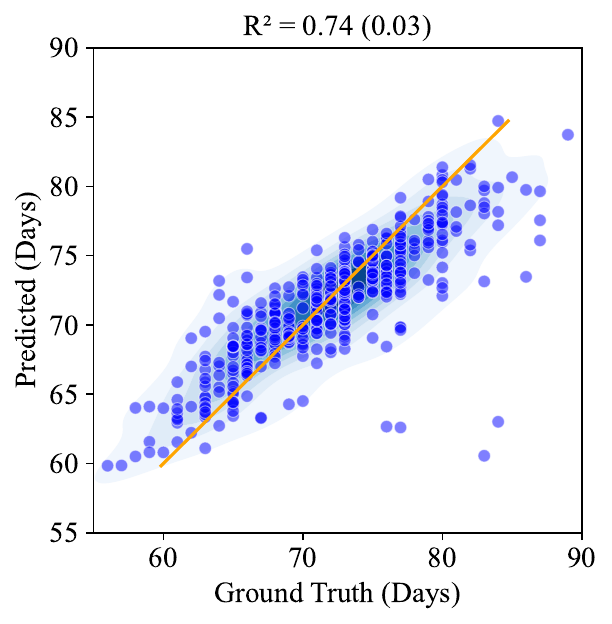} &
        \includegraphics[width=0.485\linewidth]{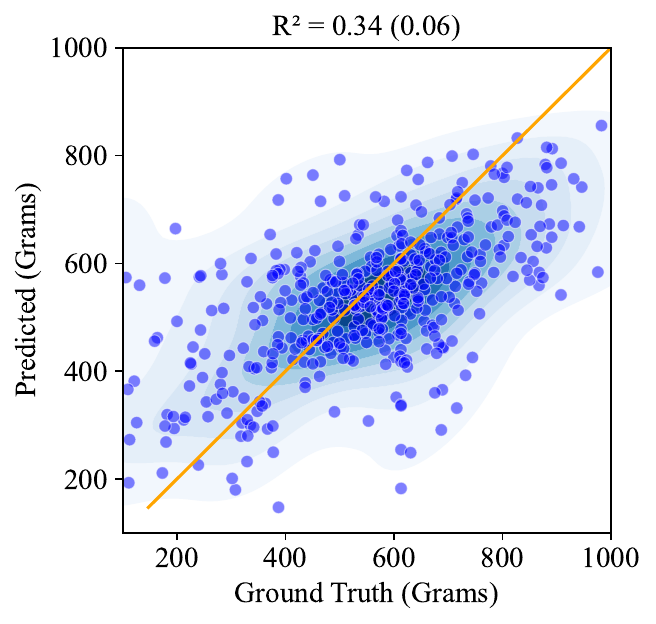} \\
        \parbox[t]{0.45\linewidth}{\centering (a) Performance on `Days to Pollen'\label{fig:CAEDaysToPollen}} &
        \parbox[t]{0.485\linewidth}{\centering (b) Performance on `Yield'\label{fig:CAETotalGrainMassGrams}}
    \end{tabular}
    \caption{Performance of CAE on `Days to Pollen' and `Yield' using Xg-Boost: (left) Days to Pollen: The x-axis indicates ground truth days (60-90 days), and the y-axis indicates predicted days (55-90 days). Points near the line y = x indicate accurate predictions. The model achieves an R² of 0.74 (±0.03), demonstrating strong prediction accuracy. (right) Yield: The x-axis indicates ground truth yield in grain mass (200-1000 grams), and the y-axis indicates predicted grain mass (200-1000 grams). Points near the line y = x indicate accurate predictions. The model achieves an R² of 0.34 (±0.06), demonstrating moderate prediction accuracy.}
    \label{fig:CAE_Results}
\end{figure}

Next, we compare this against the performance of the CAE based disentangled representation (similarly trained with multiple machine learning models out of which XgBoost yielded the best results and its performance is reported here). Figure \ref{fig:CAE_Results} shows the performance of the structured latent representation generated by the CAE. The Compositional Autoencoder (CAE) performs exceptionally well for the `Days to Pollen` trait, achieving an $r^2$ value of $0.74$. While its performance in predicting `Yield` is lower, with an $r^2$ value of $0.34$, this is unsurprising given the complexity of the genetic architecture governing yield. Accurate prediction of yield is inherently challenging due to its intricate genetic influences. Previous studies with these genotypes ~\citep{yieldpreds_measurements} involved costly and labor-intensive genotyping and manual trait measurements. These methods require significant time and effort. Considering these factors, achieving such performance using leaf hyperspectral reflectance collected only at a single time point is significant.

It is worthwhile to compare these results against recent studies based on collecting hyperspectral reflectance measurements of whole canopies instead of the leaf reflectance used here. However, we were unable to find studies reporting results on a diversity panel, so direct comparison is very difficult. The closest was work by \citet{HyperSpecImage}, who reported a $r^2 = 0.29$ and $r^2 = 0.84$ for predicting `yield' and `Days to Pollen', respectively, from hyperspectral imagery of the Genomes2Field project, which consists of around 1000 hybrids. ~\citet{MaizeYieldMachineLearning} used hyperspectral images of the canopy of a single commercial hybrid across multiple environments to predict yield with $r^2=0.33$ with a random forest model. We see that using the CAE approach on leaf scale phenotyping produces competitive results compared to state-of-the-art canopy scale phenotyping. Recent work also suggests that using the hyperspectral data to infer intermediate physiological parameters that are subsequently used to predict yield is a promising approach. For instance, ~\citet{YieldPredWaterRegimes}, used leaf reflectance and canopy reflectance to get an $r^2=0.7$ for leaf reflectance of 100 genotypes. Our findings suggest that CAE-generated latent representations hold promise for capturing relevant yield-related information. Further research is needed to explore the integration of these latent representations with other data sources to potentially improve yield prediction accuracy. 

Finally, we compared the effectiveness of using latent representations from (a) a Principal Component Analysis (PCA) on raw data, (b) latent representations from a vanilla autoencoder (AE), and (c) latent representations from a compositional autoencoder (CAE) for predicting the traits of `Days to Pollen' and `Yield'. Here, we aim to assess whether the learned latent representations offer benefits compared to using the original data directly. 

Tables \ref{table:final-yield} (yield) and \ref{table:final-DTA} (days to pollen) summarize the performance comparison using the R-squared metric (coefficient of determination) using a 5-fold cross-validation process. The tables showcase the average R-squared values (with standard deviation in parenthesis) achieved by each method and the best-performing machine learning model for that particular scenario. The performances of all the models has been given in the supplemental section.

\begin{table}[ht]
\begin{center}
\caption{A final comparison between baseline (PCA on raw data), vanilla autoencoder, and compositional autoencoder for \textbf{yield prediction}. }
\label{table:final-yield}
\begin{tabular}{ccc}
\hline
\textbf{Metric - Model} & \textbf{\begin{tabular}[c]{@{}l@{}} Avg. Values\end{tabular}} & \textbf{\begin{tabular}[c]{@{}l@{}}ML Model \end{tabular}} \\ \hline
\textbf{$R^2-CAE$} & 0.351 (0.058) & Xg-Boost Regression\\ 
\textbf{$R^2-AE$} & 0.026 (0.017) & Ridge Regression\\ 
\textbf{$R^2-PCA$} & 0.034 (0.016) & Ridge Regression\\ \hline
\end{tabular}
\end{center}
\end{table}

As observed in Table \ref{table:final-yield}, the CAE achieves a significantly higher average R-squared value (0.351) compared to both the AE (0.026) and the baseline using PCA on raw data (0.034) for predicting "Yield." This suggests that the disentangled latent representations learned by the CAE capture more relevant information for predicting yield compared to the other methods. The best performing model for all three scenarios is Xg-Boost Regression, highlighting its effectiveness for this particular regression task.

\begin{table}[ht]
\begin{center}
\caption{A final comparison between baseline (PCA on raw data), vanilla autoencoder, and compositional autoencoder for \textbf{Days to Pollen}.}
\label{table:final-DTA}
\begin{tabular}{ccc}
\hline
\textbf{Metric - Model} & \textbf{\begin{tabular}[c]{@{}l@{}} Avg. Values\end{tabular}} & \textbf{\begin{tabular}[c]{@{}l@{}}ML Model \end{tabular}} \\ \hline
\textbf{$R^2-CAE$} & 0.68 (0.034) & Xg-Boost Regression\\ 
\textbf{$R^2-AE$} & -0.01 (0.025) & Ridge Regression\\ 
\textbf{$R^2-RAW-PCA$} & 0.108 (0.02) & Ridge Regression\\ 
\textbf{$R^2-RAW$} & 0.16 (0.00) & Ridge Regression\\ \hline
\end{tabular}
\end{center}
\end{table}

Similarly, Table \ref{table:final-DTA} shows the results for predicting "Days to Pollen." Here, CAE again demonstrates a clear advantage with an average R-squared value of 0.68, significantly higher than both AE (0.106) and the baseline PCA approach (0.108). This reinforces the notion that the disentangled representations from the CAE do a better job of capturing the factors influencing the number of days to pollen in the data.

Overall, these results suggest that leveraging the latent representations learned by the CAE offers a substantial advantage for predicting both "Yield" and "Days to Pollen" compared to using the raw data directly or latent representations from the AE. This highlights the effectiveness of disentangled representations in capturing underlying factors that are relevant to these specific traits.

\clearpage

\subsection{Consistency of Latent Representations }
We evaluate the consistency of the disentangled latent representations by training the model with multiple initial conditions and evaluating its performance across different regression models. This enhances confidence in the reliability and generalizability of the learned latent representations.

The initialization of model parameters can impact the training process and the final performance of the model. Different initializations can lead to the model getting to different local minima, resulting in variable performance. To check the consistency of the performance, we trained both the networks (CAE and vanilla AE) using 4 different initial conditions. By training the model with multiple initial conditions, we can evaluate its robustness and consistency in learning informative latent representations. The tables \ref{table:CAEvsVAE-D2A} (Days to Pollen) and \ref{table:CAEvsVAE-Yield} (Yield) show a comparison of performance between a vanilla auto-encoder and compositional autoencoder for the traits of 'Days to Pollen' and 'Yield' after performing a 5-fold cross-validation. We clearly see the consistency of prediction accuracy across different model initializations.

\begin{table}[ht]
    \begin{center}
    \caption{Table shows results obtained for \textbf{Days to Pollen} trait using a vanilla autoencoder (AE) and the compositional autoencoder (CAE). The latent vectors generated using these 2 models performed differently with different ML models and the table below shows the best results among all the models that we tested.}
    \label{table:CAEvsVAE-D2A}
    
    \begin{tabular}{cccccc}
        \hline
        \textbf{Metric - Model} & \textbf{\begin{tabular}[c]{@{}c@{}}Init. 1\end{tabular}} & \textbf{\begin{tabular}[c]{@{}c@{}}Init. 2\end{tabular}} & \textbf{\begin{tabular}[c]{@{}c@{}}Init. 3\end{tabular}} & \textbf{\begin{tabular}[c]{@{}c@{}}Init. 4\end{tabular}} & \textbf{\begin{tabular}[c]{@{}c@{}}ML Model\end{tabular}} \\ \hline
        \textbf{$R^2$-CAE} & 0.681 (0.04) & 0.68 (0.035) & 0.676 (0.033) & 0.68 (0.034) & Xg-Boost Regression \\ 
        \textbf{$R^2$-AE} & 0.08 (0.02) & 0.127 (0.02) & 0.108 (0.03) & 0.110 (0.03) & Ridge Regression\\ \hline
    \end{tabular}
    \end{center}
\end{table}

\begin{table}[ht]
\begin{center}
\caption{Table shows the results obtained for \textbf{yield prediction} trait using a vanilla autoencoder (AE) and the compositional autoencoder (CAE). The latent vectors generated using these 2 models performed differently with different ML models and the table below shows the best results among all the models that we tested.}
\label{table:CAEvsVAE-Yield}
\begin{tabular}{cccccc}
\hline
\textbf{Metric - Model} & \textbf{\begin{tabular}[c]{@{}c@{}}Init. 1\end{tabular}} & \textbf{\begin{tabular}[c]{@{}c@{}}Init. 2\end{tabular}} & \textbf{\begin{tabular}[c]{@{}c@{}}Init. 3\end{tabular}} & \textbf{\begin{tabular}[c]{@{}c@{}}Init. 4\end{tabular}} & \textbf{\begin{tabular}[c]{@{}c@{}}ML Model\end{tabular}} \\ \hline
\textbf{$R^2$-CAE} & 0.351 (0.058) & 0.35 (0.054) & 0.338 (0.058) & 0.345 (0.06) & Xg-Boost Regression\\ 
\textbf{$R^2$-AE} & 0.026 (0.017) & 0.027 (0.014) & 0.029 (0.015) & 0.028 (0.015) & Ridge Regression\\ \hline
\end{tabular}
\end{center}
\end{table}
 
We finally report on varying various hyperparameters of the CAE, and their sensitivity to the downstream performance:
\begin{itemize}
    \item Masking: We evaluated the effect of input masking. Input masking improves the robustness and generalization of autoencoders by forcing them to reconstruct missing or corrupted data, which helps the model learn more significant features and patterns. This technique also acts as a regularization method, preventing overfitting and enhancing performance in various downstream tasks. Table \ref{table:MaskingP} shows the reconstruction accuracy as a function of masking fraction and suggests that 20\% masking is a good choice. We also observed that performance on the downstream task also improved by using masking while training. Table \ref{table:MaskingR2} shows $R^2$ observed for different masking percentages.  

    \begin{table}[h!]

    \centering
    \caption{CAE reconstruction accuracy for different masking \%}
    \label{table:MaskingP}
    \begin{tabular}{cc}
    \hline
    \textbf{Percentage Masking} & \textbf{Val. Loss} \\ \hline
    \textbf{$0\%$} & 0.08 \\ 
    \textbf{$20\%$} & 0.05 \\ 
    \textbf{$50\%$} & 0.05 \\ 
    \textbf{$70\%$} & 0.05 \\ \hline
    \end{tabular}
    \end{table}
    
\begin{table}[h!]
    \centering
    \caption{Downstream trait prediction accuracy ('Days to Pollen') for different masking \%.}
    \label{table:MaskingR2}
    \begin{tabular}{cc}
    \hline
    \textbf{Percentage Masking} & \textbf{$R^2$} \\ \hline
    \textbf{$0\%$} & 0.749 \\ 
    \textbf{$20\%$} & 0.757 \\ 
    \textbf{$50\%$} & 0.756 \\ 
    \textbf{$70\%$} & 0.763 \\ \hline
    \end{tabular}

\end{table}

    \item Network depth: Network depth is an important hyperparameter to explore because it directly influences the model's capacity to learn complex patterns and hierarchical representations within the data. Deeper networks can capture more intricate features and dependencies, potentially leading to improved performance on complex tasks, but they also require careful tuning to avoid issues such as vanishing gradients and overfitting. We evaluated how performance varied when the CAE network depth was varied. Table \ref{table:size_comparision} shows the performance observed for different-sized fully connected networks. We can see that the downstream performance is nearly independent of network depth. 
    \begin{table}[h!]
    \begin{center}
    \caption{Table shows the performance observed for `Days to Pollen' for different sized networks.}
    \label{table:size_comparision}
    \begin{tabular}{cccc}
    \hline
    \textbf{No. Parameters} & \textbf{\begin{tabular}[c]{@{}l@{}}\textbf{No. Layers}\end{tabular}}  & \textbf{\begin{tabular}[c]{@{}l@{}}\textbf{CAE - $R^2$}\end{tabular}}  \\ \hline
    \textbf{$14.7M$} & 4 & 0.76 \\ 
    \textbf{$5.5M$} & 3 & 0.76 \\ 
    \textbf{$2.2M$} & 2 & 0.76 \\ 
    \textbf{$392K$} & 1 & 0.76 \\ \hline
    \end{tabular}
    \end{center}
    \end{table}
    
    \item Size of the latent representation: We next evaluated how the size/dimension of the latent space affected the downstream trait prediction accuracy. Choosing a higher-dimensional latent space can result in better reconstruction accuracy; however, higher-dimensional latent spaces require larger datasets to avoid overfitting of downstream traits. This suggests a balanced approach in designing the dimensionality of the latent space to balance reconstruction accuracy (which improves with increasing latent space dimensionality) with trait regression accuracy (which improves with decreasing latent space dimensionality). 

    We remind the reader that our disentangled latent space is a vector consisting of three sets of components --- `Genotype features.' `field-level environment features,' and `plant-level environment features.' As the genotype is a common characteristic, we assign more dimensions to capture its effects. Field-level environmental features are allocated fewer dimensions, and plant-level environmental features are given the least. Table~\ref{table:LconfigSearch} shows how the performance of the downstream regression accuracy varies as the latent dimension is doubled from 10 to 20 to 40 to 80 dimensions. We see an asymptotic behavior after a latent space of 20 dimensions.   

    \begin{table}[h!]
    \begin{center}
    \caption{Table shows the performance observed for `Days to Pollen' for different latent configurations with 2.2 M training parameters.}
    \label{table:LconfigSearch}
    \begin{tabular}{cccc}
    \hline
    \textbf{Latent space dims (Geno-Env-Plant dims)} & \textbf{\begin{tabular}[c]{@{}l@{}}\textbf{CAE - $R^2$}\end{tabular}} \\ \hline
    \textbf{$10~(6-2-2)$} & 0.69 \\ 
    \textbf{$20~(12-4-4)$} & 0.76 \\ 
    \textbf{$40~(24-8-8)$} & 0.76 \\ 
    \textbf{$80~(48-16-16)$} & 0.77 \\ \hline
    \end{tabular}
    \end{center}
    \end{table}
    
\end{itemize}

\section{Conclusion}
This study introduced a novel compositional autoencoder (CAE) framework designed to disentangle genotype-specific and environment-specific features from high-dimensional data, thereby enhancing trait prediction in plant breeding and genetics programs. The CAE effectively separates these intertwined factors by leveraging a hierarchical disentanglement of latent spaces, leading to superior predictive performance for key agricultural traits such as "Days to Pollen" and "Yield." Our results demonstrate that the CAE outperforms traditional methods, including Principal Component Analysis (PCA) and vanilla autoencoders, in capturing relevant information for trait prediction.  The evaluation of various network architectures, latent space dimensions, and hyperparameter tuning further validated the robustness and generalizability of the CAE model. Specifically, the CAE showed consistent performance improvements across different initialization conditions and regression models, underscoring its reliability in practical applications. 

By effectively disentangling genotype and environment-specific features, the CAE offers a powerful tool for improving the accuracy and reliability of predictive models in agriculture, ultimately contributing to more informed decision-making in breeding programs and agricultural management. There are several avenues for future work. First, it will be interesting to explore the viability of compositional autoencoders for making trait predictions using the disentangled GXE features using other sensing modalities~\cite{shrestha2024plot} like (a) UAV-based hyperspectral imagery and (b) satellite-based multispectral imagery. Second, applying CAE to time-series high-dimensional data collected on diversity panels can produce disentangled low-dimensional time trajectories that could provide biological insight. Finally, integrating these disentangled latent representations with other data (crop models, physiological measurements) may be a promising approach for creating accurate end-of-season trait prediction models using mid-season data.

We conclude by identifying the following limitations of our work: (a) We evaluated the performance of the CAE on two specific traits that were phenotyped in the field experiments. Our future work will focus on evaluating the CAE on a broader range of traits; (b) Our study is based on hyperspectral reflectance data from a specific maize diversity panel. Our future work is focused on extending this to other datasets and environments; (c) While we demonstrate the technical advantages of disentanglement, it is not immediately clear how to connect these disentangled features to biological insights.

\section*{Conflict of Interest Statement}
The authors declare that the research was conducted in the absence of any commercial or financial relationships that could be construed as a potential conflict of interest.

\section*{Author Contributions}

T.Z.J and B.G conceived the approach. M.C.T and J.C.S led field experiments and data collection. A.P and T.Z.J developed the machine learning pipeline. A.P and M.C.T performed computational experiments and analysis. All authors participated in the project implementation and completion. All authors contributed to the final manuscript production. 

\section*{Funding}
This work was supported by the AI Institute for Resilient Agriculture (USDA-NIFA 2021-67021-35329) and Iowa State University Plant Science Institute.

%\section*{Acknowledgments}
%This work was supported by the AI Institute for Resilient Agriculture (USDA-NIFA 2021-67021-35329) and Iowa State University Plant Science Institute.

\section*{Data Availability Statement}
Hyperspectral reflectance data collected in 2020 and 2021 are both deposited in  
\href{https://doi.org/10.6084/m9.figshare.24808491.v4}{FigShare}. All code is available at \href{https://bitbucket.org/baskargroup/cae_hyperspectral/src/main/}{bitbucket}.

%\section*{Declaration of generative AI and AI-assisted technologies in the writing process}

%During the preparation of this work the author(s) used ChatGPT in order to improve readability. After using this tool/service, the author(s) reviewed and edited the content as needed and take(s) full responsibility for the content of the published article.

\bibliographystyle{Frontiers-Harvard} %  Many Frontiers journals use the Harvard referencing system (Author-date), to find the style and resources for the journal you are submitting to: https://zendesk.frontiersin.org/hc/en-us/articles/360017860337-Frontiers-Reference-Styles-by-Journal. For Humanities and Social Sciences articles please include page numbers in the in-text citations 

\bibliography{frontiers}

\clearpage  % Start a new page for the supplementary section
\appendix   % Switch to appendix mode

\section{Supplementary Materials}
\beginsupplement
\subsection{Data Normalization}
\begin{figure}
    \centering
    \includegraphics[width=1\linewidth]{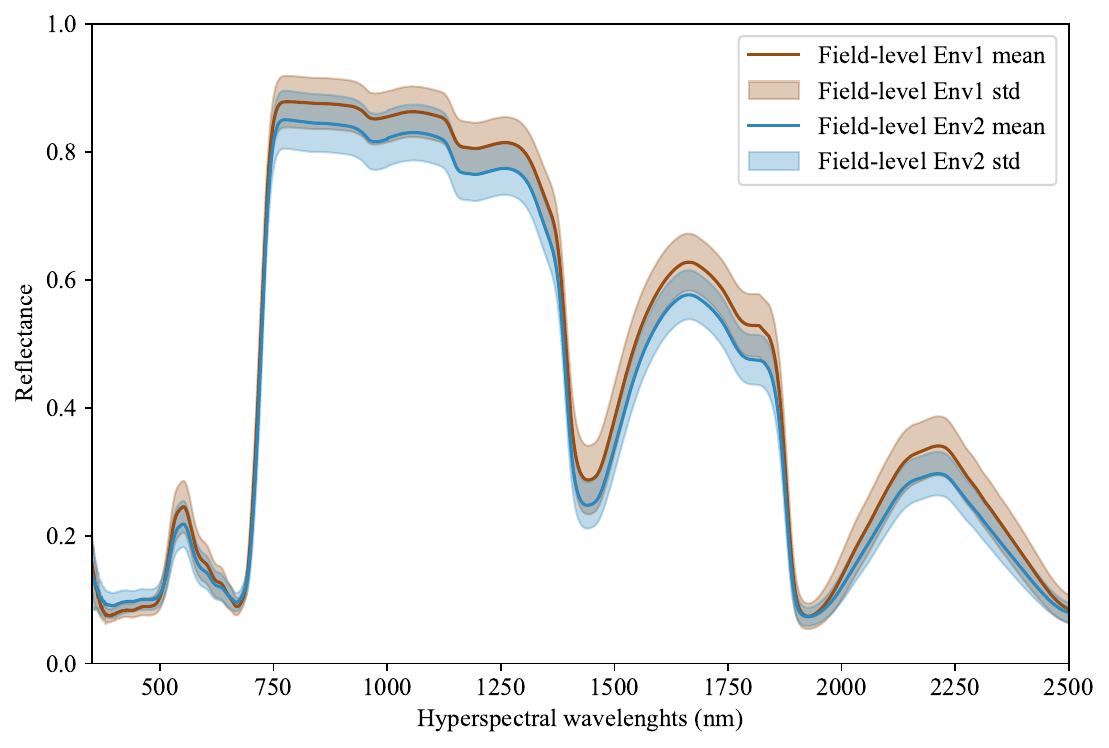}
    \caption{Hyperspectral Data after normalization}
    \label{fig:Data_norm}
\end{figure}

We use Min-Max normalization to normalize the data. 
Figure \ref{fig:Data_norm} is the data visualization after normalization.
\subsection{Neural Network Training}

A neural network is a mathematical model inspired by the structure and function of biological neural networks. It consists of interconnected units or nodes called neurons, organized into layers. The most basic type of neural network is the feedforward neural network, where information moves in only one direction, forward, from the input nodes, through the hidden nodes (if any), and to the output nodes.

\subsubsection*{Mathematical Representation:}

\begin{enumerate}
    \item \textbf{Input Layer:} The input layer consists of input neurons that receive the input features (denoted as $x_1, x_2, \ldots, x_n$).

    \item \textbf{Hidden Layers:} Each neuron in the hidden layers applies a weighted sum to its inputs and then passes it through an activation function. The output of the $j$-th neuron in the $k$-th layer can be expressed as:
    \[
    a_j^{(k)} = \sigma\left(\sum_{i=1}^{n} w_{ij}^{(k)} x_i + b_j^{(k)}\right)
    \]
    where $w_{ij}^{(k)}$ is the weight associated with the connection between the $i$-th neuron in the $(k-1)$-th layer and the $j$-th neuron in the $k$-th layer, $b_j^{(k)}$ is the bias term for the $j$-th neuron in the $k$-th layer, and $\sigma$ is the activation function (e.g., sigmoid, ReLU).

    \item \textbf{Output Layer:} The output layer produces the final output of the network. The process is similar to that of the hidden layers, but the output might pass through a different activation function suitable for the specific task (e.g., softmax for classification).
\end{enumerate}

\subsubsection*{Training the Neural Network:}

The training of a neural network involves adjusting the weights and biases to minimize the difference between the predicted output and the actual output. This process is typically done using the backpropagation algorithm and an optimization technique like gradient descent.

\begin{enumerate}
    \item \textbf{Loss Function:} A loss function ($L$) measures the difference between the predicted output ($\hat{y}$) and the actual output ($y$). Common loss functions include mean squared error for regression tasks and cross-entropy loss for classification tasks.

    \item \textbf{Backpropagation:} This algorithm computes the gradient of the loss function with respect to each weight and bias in the network by applying the chain rule of calculus. It starts from the output layer and propagates the error backward through the network. The gradient of the loss with respect to the weights in layer $k$ is given by:
    \[
    \frac{\partial L}{\partial w_{ij}^{(k)}} = \frac{\partial L}{\partial a_j^{(k)}} \cdot \frac{\partial a_j^{(k)}}{\partial w_{ij}^{(k)}}
    \]
    where $\frac{\partial L}{\partial a_j^{(k)}}$ is the error propagated from the next layer, and $\frac{\partial a_j^{(k)}}{\partial w_{ij}^{(k)}}$ is the derivative of the activation function with respect to the weights.

    \item \textbf{Optimizer:} An optimizer uses the gradients calculated by backpropagation to update the weights and biases. The L-BFGS optimizer is a quasi-Newton method that approximates the Hessian matrix to guide the search for the minimum. The weight update equation for L-BFGS can be expressed as:
    \[
    w^{(k+1)} = w^{(k)} - \alpha H_k \nabla L(w^{(k)})
    \]
    where $w^{(k)}$ are the weights at iteration $k$, $\alpha$ is the step size, $H_k$ is the approximate inverse Hessian matrix, and $\nabla L(w^{(k)})$ is the gradient of the loss function at iteration $k$.
\end{enumerate}

This iterative process of forward pass, computing loss, backward pass, and updating weights is continued until the model is sufficiently trained. After training, the neural network can be used for predictions or further analysis.
\subsection{Optimizer}
\subsubsection{LBFGS}
To train this neural network, we used the LBFGS optimizer.
In the case of a neural network, the optimizer plays a very important role. It drives the weights of the network to a point in the $N$-dimensional space ($N$ is the number of parameters of the neural network) for which, the loss function for this network is at its minimum. 
LBFGS is a quasi-newton method of optimization that approximates the hessian instead of calculating it at every iteration to reduce the complexity and time. 
Newton's method is used to find the minima of a non-quadratic function. It typically starts with a random point and approximates a quadratic around that point and finds the minima for that quadratic and repeats these steps until a minima for the non-quadratic is found. Quadratic is approximated using the `Taylor series' expansion. \begin{equation}
    g(x) = f(a) + f'(a)(x-a) + \frac{f''(a)}{2!}(x-a)^2 + \frac{f'''(a)}{3!}(x-a)^3 + \cdots
\end{equation}
where:
\begin{itemize}
    \item \( f^{(n)}(a) \) denotes the \( n \)-th derivative of \( f \) evaluated at the point \( a \),
    \item \( n! \) is the factorial of \( n \).
\end{itemize}

\subsubsection*{Hessian Approximation in L-BFGS:}

L-BFGS maintains a history of the last $m$ updates of the position vectors ($s_k$) and the gradient vectors ($y_k$), where $k$ indexes the iteration. The position and gradient vectors are defined as:
\[
s_k = w^{(k+1)} - w^{(k)}, \quad y_k = \nabla L(w^{(k+1)}) - \nabla L(w^{(k)})
\]

The approximate inverse Hessian matrix ($H_k$) is updated at each iteration using this history. The update formula is derived from the BFGS update formula but modified to use limited memory. The approximation starts with an initial estimate $H_k^0$, which is often chosen as a scaled identity matrix:
\[
H_k^0 = \gamma_k I
\]
where $\gamma_k$ is a scaling factor that can be computed using different strategies, one common choice being:
\[
\gamma_k = \frac{s_{k-1}^\top y_{k-1}}{y_{k-1}^\top y_{k-1}}
\]

Then, the approximate inverse Hessian $H_k$ is updated using the formula:
\[
H_k = \left( V_k^\top H_k^0 V_k + \rho_k s_k s_k^\top \right)
\]
where $V_k = I - \rho_k y_k s_k^\top$ and $\rho_k = \frac{1}{y_k^\top s_k}$.

\subsubsection*{Step Size Determination in L-BFGS:}

The step size ($\alpha_k$) in L-BFGS is typically determined using a line search method that satisfies the Wolfe conditions. The line search aims to find a step size that ensures a sufficient decrease in the loss function and a sufficient slope of the gradient. The Wolfe conditions are:

\begin{enumerate}
    \item \textit{Sufficient Decrease Condition (Armijo Condition):}
    \[
    L(w^{(k)} + \alpha_k p_k) \leq L(w^{(k)}) + c_1 \alpha_k \nabla L(w^{(k)})^\top p_k
    \]

    \item \textit{Curvature Condition:}
    \[
    \nabla L(w^{(k)} + \alpha_k p_k)^\top p_k \geq c_2 \nabla L(w^{(k)})^\top p_k
    \]
\end{enumerate}
where $0 < c_1 < c_2 < 1$ are constants, $p_k = -H_k \nabla L(w^{(k)})$ is the search direction, and $\alpha_k$ is the step size.

The line search algorithm iteratively adjusts $\alpha_k$ until the Wolfe conditions are satisfied, ensuring that the step size leads to a sufficient decrease in the loss function while maintaining the curvature condition.

\subsection{Parameter Exploration}
\begin{table}[ht]
\begin{center}
\caption{Table shows the performance observed for different masking percentages. Val. Loss = Coereff Loss + L2 Loss.}
\label{table:MaksingP}
\begin{tabular}{|c|c|c|c|}
\hline
\textbf{Percentage Masking} & \textbf{\begin{tabular}[c]{@{}l@{}}\textbf{Val. Loss}\end{tabular}} \\ \hline
\textbf{$0\%$} & 0.08 \\ \hline
\textbf{$20\%$} & 0.05 \\ \hline
\textbf{$50\%$} & 0.05 \\ \hline
\textbf{$70\%$} & 0.05 \\ \hline
\end{tabular}
\end{center}
\end{table}
Autoencoders are often trained with many different pre-text tasks. A very prevalent method is masking. During the training, the input is partially masked (with zeros) and fed into the Autoencoder, and the reconstruction loss is calculated with respect to the original input data. This forces the network to learn better. The table \ref{table:MaksingP} shows the results with different masking percentages.

\subsection{Downstream Models}
\subsubsection{Random Forests}
Random Forest is an ensemble method composed of numerous decision trees. For regression tasks, it averages the outputs of these trees to obtain the final output, while for classification tasks, it typically uses majority voting. Decision trees recursively split the data into subsets based on attribute tests that maximize information gain. Information gained in each split can be calculated using either entropy or the Gini index. The formula for entropy is $Entropy = -\sum_{i=1}^{n} P(x_i) \log_b P(x_i)$, where $b$ is usually 2 representing binary logarithm. The formula for gini index is $Gini Index = 1- \sum{p_i}^2$, where $p_i$ is the probability of class $i$. The table \ref{table:Parameters} presents the parameters used for Random Forests in the downstream models.
\subsubsection{XGBoost}
XGBoost (boosting method) is also an ensemble method. It builds trees sequentially, where each tree tries to correct the errors made by the previous ones. It focuses on boosting weak learners (typically shallow trees) by focusing more on the instances that were misclassified or had higher errors by the earlier trees.

\begin{table}[htbp]
\centering
\caption{Performance Comparison of Models on Days to Pollen (R-squared)}
\label{table:Parameters}
\begin{tabular}{lllc}
\toprule
\textbf{Feature Extraction Model} & \textbf{Prediction Model} & \textbf{Parameters} & \textbf{R-squared} \\
\midrule
CAE (Ours) & XGBoost & \begin{tabular}[c]{@{}l@{}}max depth: 15,\\ n estimators: 1500,\\ learning rate: 0.01\end{tabular} & \textbf{0.76} \\
           & PLSR    & 10 components                                            & 0.02  \\
           & Ridge   & alpha = 0.001                                            & 0.027 \\
           & Random Forest & n estimators = 300                                 & 0.753 \\
\midrule
AE & XGBoost & \begin{tabular}[c]{@{}l@{}}max depth: 1,\\ n estimators: 1500,\\ learning rate: 0.01\end{tabular} & -0.01 \\
           & PLSR    & 15 components                                            & -0.012 \\
           & Ridge   & alpha = 0.001                                             & -0.01 \\
           & Random Forest & n estimators = 300                                 & -0.03  \\
\midrule
Raw Reflectance & Ridge & alpha = 0.05                                           & \textbf{0.16}  \\
                & PLSR  & 15 components                                          & 0.148 \\
                & Random Forest & n estimators = 300                             & 0.08  \\
                & XGBoost & \begin{tabular}[c]{@{}l@{}}max depth: 3,\\ n estimators: 1000,\\ learning rate: 0.01\end{tabular} & 0.09  \\
\bottomrule
\end{tabular}
\end{table}

\subsubsection{Ridge Regression}
This works in the same fashion as a typical least square regression where, the algorithm aims the minimize the sum of squared residuals. The change here is the addition of a regularization term that reduces the variance, thereby improving the performance of the model for data that suffers from multicollinearity. 
\begin{equation}
    J(\beta) = \sum_{i=1}^{n} (y_i - \beta_0 - \sum_{j=1}^{p} X_{ij} \beta_j)^2 + \lambda \sum_{j=1}^{p} \beta_j^2
\end{equation}
where:
\begin{itemize}
    \item \( y_i \) is the observed output,
    \item \( \beta_0 \) is the intercept term,
    \item \( X_{ij} \) are the predictor variables,
    \item \( \beta_j \) are the regression coefficients,
    \item \( \lambda \) is the regularization parameter,
    \item \( n \) is the number of observations, and
    \item \( p \) is the number of predictor variables.
\end{itemize}

\end{document}